# **P**redictive **R**adiomics for **E**valuation of **C**ancer **I**mmune **S**ignatur**E** in **G**lio**b**lasto**m**a: the **PRECISE-GBM study**

Prajwal Ghimire[1,2], Junjie Li[3], Liu Yaou[3], Marc Modat[1], Thomas Booth[1,4]

1. School of Biomedical Engineering & Imaging Sciences, King's College London, UK
2. Department of Neurosurgery, King's College Hospital, London, UK
3. Department of Neuroradiology, Beijing Tiantan Hospital, Beijing, China
4. Department of Neuroradiology, King's College Hospital, London, UK

**Corresponding Author:**

Dr Thomas Booth

School of Biomedical Engineering and Imaging Sciences, Kings College London

Email: thomas.booth@kcl.ac.uk

**Word Count:**

Abstract : 226; Importance of study: 109; Manuscript: 5690 (excluding references)

Figures: 4, Tables: 2

Supplemental File: 1

**Abstract**

**Background**

Radiogenomics allows identification of radiological biomarkers for genomic phenotypes. In glioblastoma, these biomarkers could potentially complement patient stratification strategies. We aim to develop and analytically validate radiological biomarkers that capture immune cell signatures within IDH-wildtype glioblastoma microenvironment using radiogenomic analysis.

**Methods**

This was a retrospective multicenter study using curated open-access anonymized imaging and genomic data from TCGA-GBM, CPTAC, IvyGAP, REMBRANDT and CGGA datasets. Imaging data consisted of MRI-based radiomic features extracted from necrotic core, enhancing and edema regions of deep learning-based auto-segmented tumors. Radiomic feature selections were performed using nested cross-validated LASSO. Support vector machine and ensemble models were trained using seventeen immune and cell-specific score labels extracted from deconvoluted transcriptomic data using pan-cancer and glioblastoma immune signature matrices as reference standards. Seventeen classifier models trained in three cross-cohort strategies were validated on three held-out datasets assessing stability and generalizability.

**Results**

One-hundred-and-seventy-six patients were included in the study. The immune-related radiomic signatures obtained after feature selection were shape, first order and higher order radiomic features. Models predicting macrophage subtype immune signature showed stable mean performance on balanced accuracy (0.67) and precision (0.89) metrics for three independent holdout datasets with ensemble model outperforming support vector machine model.

**Conclusion**

Radiogenomic models non-invasively predicted the macrophage subtype M0 immune signature in IDH-wildtype glioblastoma. These biomarkers have the potential to stratify patients for immunotherapy within prospective glioblastoma clinical trials.

**Key Points**

- Radiogenomic models predicted macrophage M0 status in glioblastoma.
- These biomarkers have the potential to stratify patients for immunotherapy trials.

**Importance of the study**

Radiogenomic analysis provides a promising complementary avenue for non-invasive profiling of this immune TME pre-operatively for trial enrichment. For the first time, the PRECISE-GBM study used a radiogenomic framework to classify immune cell profiles for IDH-wildtype glioblastoma using the largest known open-source dataset of matched MR images and transcriptome. Biomarkers for macrophage subtype M0 gave good performance accuracy in holdout test sets and the models were stable. In summary, we developed and analytically validated an MRI-based pipeline for interrogating glioblastoma immunobiology for macrophage M0 subtype non-invasively which has the potential for clinical trial enrichment. The pipeline architecture and rigorous methods of evaluation can be used in other radiogenomic studies.

## 1. Introduction:

The immune tumor microenvironment (TME) in IDH-wildtype (IDH-wt) glioblastoma[1-10] is immunosuppressive compared to other cancers including melanoma and lung cancer with significant inter- and intra-tumoral genetic heterogeneity.

There is a potential role for immunotherapy personalized to patients' immune TME and influencing survival of these patients, albeit overall evidence is currently limited with a preponderance of pre-clinical studies [1-7,16-22,24,25,28]. The current standard of care for identifying immune biomarkers is based on tumor tissue biopsy, histopathology and immunohistochemistry[10]. While these approaches provide detailed molecular and cellular insights, radiogenomic analysis using pre-operative MRI offers a complementary non-invasive approach that may help characterize immune related tumor features for clinical trial enrichment prior to biopsy for neo-adjuvant trial stratification[8,23-25,29-31]. It is plausible that enriched trials are needed in this setting as whilst there are promising mechanisms of action, clinical efficacy of neo-adjuvant treatment remains to be proven[23]. Potentially, these radiogenomic approaches can also be useful for inoperable patients or deep-seated lesions, where tissue analysis is not possible, complementing clinical judgement for treatment choices."

Recently, tissue-based immune signatures have been developed using bioinformatics-based computational tools which input these bulk tumor transcriptomic data to produce estimates of immune cells within the TME [32]. Computational advances have also led to the recent expansion of radiogenomics, which allows underlying genomic information to be represented as features derived from pixel-based image data. Here, machine learning can be leveraged to regress or classify these features, and such approaches have been applied to MR images of glioblastoma[33-37].

We hypothesized that such computational advances would allow the development of diagnostic biomarkers from pre-operative MRI images, that can predict the tumor's immune status at the time of provisional glioblastoma diagnosis.

Therefore, we aimed to use pre-operative MR images to identify immune signatures derived from transcriptomic data, by developing and analytically validating[38] a radiogenomic multi-modality machine learning architecture. To achieve this aim, an initial objective was to curate and harmonize the largest open-access IDH-wt glioblastoma dataset containing matched MRI and transcriptomic information, which would provide sufficient data for training and holdout external testing.

Box 1: Immune TME of IDH-wildtype glioblastoma and implications for immunotherapy

- **Immunotherapy background.** IDH-wt glioblastoma is poorly responsive to immunotherapy with only a small subgroup of hypermutated tumors - associated with mismatch repair deficiency and high neoantigen load that activates T cells - showing efficacy in recent studies[7-9,11-14]. Nonetheless, pre-clinical studies have suggested that the glioblastoma immune axis, including targets in T cells, tumor-associated macrophages and antigen-presenting cells, are still relevant as therapeutic targets[15-22].

- **TME.** Constraints in trial enrichment, tumor heterogeneity and an immunosuppressive type of TME, are the likely reasons for failure in most of these immunotherapy clinical trials[1-3,7,14]. This TME includes immune cells such as CD4, CD8 T cells, T regulatory cells (Treg), dendritic cells (DCs), natural killer cells (NK cells), microglia cells and tumor-associated macrophages (TAMs) including M0, M1 and M2 macrophages – all of which play an important role in determining potential response to immunotherapy treatment[1-3,12,15-22,26,27].

- **Clinical efficacy.** In a study in 2025[8], neo-adjuvant immunotherapy in newly diagnosed IDH-wildtype glioblastoma demonstrated marked immune activation including tumor-infiltrating lymphocytes that induced an interferon response within the TME. Cell cycle-related gene expression within tumor cells was downregulated leading to a treatment response. This has led to the ongoing phase 2 clinical trial (GIANT) with neo-adjuvant followed by adjuvant immunotherapy in newly diagnosed IDH-wildtype glioblastoma[23]. In a randomized clinical trial in 2019[24], neo-adjuvant immunotherapy in recurrent, surgically-resectable glioblastoma demonstrated enhancement of local and systemic anti-tumor immune responses. The enhancement manifest as upregulation of T cell-related gene expression and augmented clonal expansion of T cells that act on the tumor cells producing anti-tumor activity and led to a survival benefit compared to adjuvant-only treatment. Similarly, in another clinical trial in 2024[25] examining recurrent, surgically-resectable glioblastoma, addition of neoadjuvant immunotherapy was associated with downregulation of cancer proliferation genes and upregulation of T cell gene expression leading to anti-tumor activity. The genetic changes were positive risk factors for survival.

- **Pre-clinical efficacy.** Cytotoxic immune populations such as CD8 T cells, NK cells, and DCs promote anti-tumor immunity by mediating tumor cell killing and antigen presentation, whereas immunosuppressive populations including T regulatory cells, TAMs (especially M2 macrophages), and microglia inhibit effective immune activation by suppressing effector T cells, PD-L1 expression and tumor supportive inflammation[16-22]. Experimental studies have shown that glioma-induced PD-L1 expression on TAMs suppresses T-cell activity, contributing to immune evasion[17]. Conversely, activation of DCs and expansion of CD8 T cells following checkpoint blockage or radiation therapy can generate durable anti-tumor responses in preclinical glioma models[16,18,19]. The studies also demonstrated that checkpoint blockade responses depend on immune cell composition with response seen only in immunocompetent models with active cellular response.

## 2. Materials and Methods

We performed a biomarker development and analytical validation study[38] using retrospective international open-access multicenter data (**Fig. 1a**) for newly diagnosed IDH-wt glioblastoma as per WHO 2021 classification[10]. The study reporting followed the Checklist for Artificial Intelligence in Medical Imaging (CLAIM) and TRIPOD-AI checklist (**Supplemental Table S6,7**).

### 2.1 Datasets

We curated and harmonized matched data consisting of pre-operative MRI images and transcriptomic data derived from tumor tissue samples (**Fig. 1a**). Matched datasets obtained from open access data portals (**Supplemental Fig. S1**) were The Cancer Genome Atlas[39] (TCGA-GBM), Clinical Proteomic Tumor Analysis Consortium[40] (CPTAC), Ivy Glioblastoma Atlas Project (IvyGAP)[41], REpository for Molecular BRAin Neoplasia DaTa[42] (REMBRANDT) and the Chinese Glioma Genome Atlas (CGGA)[43].

Diagnostic pre-operative $T_1$-weighted (T1), $T_2$-weighted (T2), $T_1$-weighted contrast-enhanced (T1 CE) and $T_2$ Fluid Attenuated Inversion Recovery (FLAIR) MRI sequences were obtained and all 4 sequences used in our pipeline. The CGGA MRI dataset contained 6 sets of images including pre and post contrast T1, T2 and FLAIR sequences, but unlike other datasets, it also contained 66 patients without FLAIR sequences. Because we wanted to include this large, matched dataset, we generated synthetic FLAIR images using a pix2pix-based[44] generative adversarial network (GAN) model (**Fig. 1b; Supplemental Table S5**). To guarantee the veracity of holdout test sets, all image sets containing synthetic images were used for training only.

Three combinations of training and testing data were curated (**Fig. 1a**). Group 1 consisted of TCGA, CPTAC, CGGA, REMBRANDT training data with IvyGAP used as a holdout test dataset; Group 2 trained on CPTAC, IvyGAP, CGGA, REMBRANDT and tested on TCGA; and Group 3 trained on TCGA, IvyGAP, CGGA, REMBRANDT and tested on CPTAC.

REMBRANDT and CGGA datasets were not used as holdout datasets as REMBRANDT contained microarray data (which is becoming obsolete due to the superior resolution and flexibility of RNA sequencing) and, as described above, CGGA contained synthetic MRI data.

As well as genomic information, we also collected clinical information regarding patient demographics and survival data. We also calculated tumor volume.

## 2.2 Transcriptomic data and construction of classifier labels

RNA sequencing data (TCGA-GBM, CPTAC, IvyGAP, CGGA) and microarray data (REMBRANDT) for IDH-wt glioblastoma[10] were collected. REMBRANDT data contained normalized micro-array gene expression data. TCGA-GBM and CPTAC data contained unstranded fragment per kilobase per million mapped reads (fpkm) values for each patient, CGGA data contained normalized fpkm values, and IvyGAP data contained gene-level fpkm values. The unstranded fpkm values were converted to stranded fpkm values (gffutils v0.12, https:/github.com/daler/gffutils) which were then converted to gene-level fpkm values by calculating the sum of fpkm values over all genes. Following this step, all the datasets were harmonized (combat v0.3.3, epigenelabs.github.io/pyComBat/). We included REMBRANDT microarray data, which differs from the sequencing data, to increase the training dataset size whilst potentially increasing model generalizability for bulk gene expression data.

The digital cytometry tool CIBERSORTx[45] (Stanford University, USA) was used to estimate cell scores from transcriptomic data through a deconvolution algorithm which allowed us to use both a pan cancer signature matrix and a glioblastoma signature matrix[46]. For each matrix, the immune score (i.e., absolute overall immune score) as well as cell-specific scores (overall T cells, TAMs (overall macrophages), DC, NK cell scores) were extracted. Additional cell-specific scores were extracted from the pan cancer signature matrix (CD8 T cells, CD4 T cells, Treg cells, and M0, M1 and M2 macrophage scores) and the glioblastoma signature matrix (microglia). The scores were divided into binary labels (0: low score, 1: high score) using a gaussian mixture model[47,48] (GMM) (Gaussian Mixture package, Scikit-learn v1.2.2, scikit-learn.org/stable/install.html). These 17 labels were used as classes for training and testing of classifiers.

### 2.3 Radiomic extraction and feature selection

The images of each patient underwent automated registration, skull stripping and segmentation of the following tumor regions: necrotic core, enhancing tumor and "edema", using v2.1.0 DeepBraTumIA[49,50] (University of Bern, Switzerland). Pyradiomics[51] (Harvard University, USA), incorporated within our pipeline, was applied to the segmented tumor for each sequence (per segmented region across all sequences) and radiomic features were extracted (**Supplemental Fig. S2**). Examples of auto-segmented tumor regions of these glioblastoma patients are demonstrated in **Supplemental Fig. S6**.

Principal component analysis (PCA package, scikit-learn v1.2.2, scikit-learn.org/stable/install.html) was performed on all radiomic data, and Hotelling's T2 ellipse employed (matplotlib v3.9, matplotlib.org/stable/) to understand, then remove outliers. Outliers removed were both Toshiba MRI scans from 2012 with unknown Tesla strength (CPTAC C3L-01146, C3L-01156) and one 1.5 T MRI scan from 2000 with an unknown manufacturer (TCGA-06-0168) which fell outside the 99% threshold of the Hotelling's T2 ellipse (**Supplemental Fig. S3**).

For each of the three Group training datasets, the radiomic features were harmonized using neuroCombat (v0.2.12, github.com/Jfortin1/neuroCombat) to remove batch effects thereby reducing batch-driven spurious variation[52]. Prior to training, the harmonized radiomic data from the three Group training datasets each underwent Z score normalization to reduce the untoward influence of higher radiomic values during model training (**Supplemental Fig. S4**). We note here that following these harmonization and normalization steps of each Group training dataset, we extracted a function to be fixed within each respective Group pipeline to allow unseen non-harmonized and non-normalized single or batch patient data from holdout test sets to be transformed identically to the training data. This pipeline transformation strategy optimized unseen data for classification whilst preventing pre-processing data leakage.

To reduce the number of features for classification, the most discriminant features within each of the three Group training datasets were obtained through feature selection using Least Absolute Shrinkage and Selection Operator (LASSO) separately, with nested 5-fold cross validation (LASSO-CV) (LassoCV, scikit-learn v1.2.2, scikit-learn.org/stable/install.html) applied for leakage-free feature selection.

**2.4 Training and validation of the models**

Support vector machine (SVM) models were central to our subsequent classifier strategy. We trained SVM alone as well as ensembled with influence from random forest (RF) and histogram-based gradient boost (GB) using an arbitrary voting classifier 2:1:1 weight split (SVM:RF:GB)[53-57]. Classifier performances were compared for each class label.

For both SVM and ensemble models, we performed an initial cross validation step to ensure stability and robustness using training data alone. The initial step consisted of 5-fold CV in each of three Group training datasets, stratified by class to mitigate imbalance, to select optimal hyperparameters and to select the subsequent best trained model for the respective Group (trained on 80% of the training data for the respective Group, because 5-fold CV). The best model with tuned hyperparameters was then re-trained on 100% of the training data from that Group to give the final fixed model. Each final fixed model was then tested on the respective holdout dataset for that particular Group (IvyGAP, TCGA or CPTAC) (**Supplemental Fig. S5**). There was no holdout dataset leakage as only training data was used to build the respective final fixed models. In summary, three dataset-specific holdout predictions were obtained for IvyGAP, TCGA and CPTAC holdout test sets using the respective final fixed models (**Fig. 1c**).

**2.5 Statistics**

We applied descriptive and comparative statistics to Group dataset baseline characteristics. ANOVA was utilized for group-wise comparison of age at diagnosis, overall survival (OS) and tumor volume,  and Chi-square test for gender. A log rank test was also performed for comparison of OS among the datasets. Calibration analyses and Shapley Additive exPlanations (SHAP) were utilized for interpretability of model stability and explainability.

The performance accuracy of the three independent experiments was assessed in two ways.

First, three comprehensive standalone holdout test results were calculated to demonstrate generalizability of the models for each immune label (precision, accuracy, balanced accuracy, recall, F1 score and Matthews correlation coefficient-MCC). To summarize the results in a relevant and conservative manner, we focused on precision, balanced accuracy and MCC. Precision was chosen as immunotherapy drug trials require enrichment of patients with a favorable immune profile and a low false positive rate. Balanced accuracy was important as the immune label classes were imbalanced. In addition, MCC, which evaluates correlations between observed and expected predictions, was employed to give a conservative balanced measure of classification quality despite class imbalance[58].

Second, we determined if performance accuracy was superior in one model over another for a particular immune category by leveraging the results of the three independent experiments. We performed bootstrapped resampling[59,60] with 10,000 stratified resamples of the held-out test predictions generated to estimate the distribution of mean metric differences between model pairs (e.g., "SVM for pan cancer signature matrix" vs "ensemble for glioblastoma signature matrix"), along with 95% confidence intervals (CIs). Statistical significance was assessed using raw CI-based p value, with Bonferroni corrections applied for multiple comparisons. A p-value $< 0.05$ was considered statistically significant prior to multiple comparison tests being applied.

Models that classified the immune category accurately on the basis of precision, balanced accuracy and MCC - supplemented by bootstrapping results to determine the best model within that category- were selected as validated models.

### 2.6 Processing

The training of radiogenomic models was performed using a 12th Gen Intel(R) Core (TM) i7-1260P processor; segmentation and generation of synthetic images were performed using an NVIDIA GeForce RTX 3080 GPU processor.

## 3. Results

One-hundred-and-seventy-six patients with IDH-wt glioblastoma with matched transcriptome and imaging data were included in the study. There was no significant difference in overall survival (OS) when all datasets were compared (**Supplemental Fig. S7**), nor when datasets were combined into the three Groups across age, gender, OS and tumor volume (**Table 1**).

### 3.1 Radiomics

Following feature selection with LASSO-CV using the harmonized radiomic data within each training group 1-3, a set of radiomic features were obtained for each binary transcriptome-derived immune label category. For visualization, all these features from three groups of models were encompassed together, duplicates were removed and unique features for each immune category were compiled. The selected feature sets were made up from 2069 distinct radiomic features which were a combination of shape, first order and higher order (GLSZM: grey-level size zone matrix; GLCM: grey-level co-occurrence matrix; GLDM: grey-level difference matrix; NGTDM: neighborhood grey tone difference matrix; GLRLM: grey-level run length matrix) features. There was a similar contribution in the number of selected features from the different segmented regions with necrotic core contributing 701 (33.9%), enhancing tumor contributing 643 (31.1%), and edema region contributing 725 (35%) features. Likewise, there was a similar contribution in the number of selected features from the different MRI sequences with T1 contributing 534 (25.8%), FLAIR contributing 496 (24%), T2 contributing 491 (23.7%) and T1CE contributing 548 (26.5%) features. All immune label categories incorporated 163-179 selected features: overall immune score was represented by 179 (8.6%), CD4 T cells by 175 (8.5%), CD8 T cells by 171 (8.3%), T regulatory cells by 171 (8.3%), overall T cells by 178 (8.6%), TAMs by 178 (8.6%), M0 macrophages by 171 (8.3%), M1 macrophages by 163 (7.8%), M2 macrophages by 177 (8.5%), microglia by 168 (8.1%), NK cells by 173 (8.4%) and DC by 165 (8%) unique radiomic features **(Fig. 2, Supplemental Table S1)**.

### 3.2 Model Results

Three sets of experiments were performed using three different combinations of training and holdout test datasets, referred to as Group 1, 2 and 3 (**Table 2**). Results were compared according to the holdout datasets, type of model deployed (SVM or ensemble), and the signature matrix (pan cancer or glioblastoma signature matrices) underlying the construction of 17 binary immune labels (**Fig. 3**). Five immune label categories were shared between the two matrices, microglia were specific to glioblastoma-specific matrix, and 6 categories (CD4, CD8 T cells, T reg cells and macrophage subtypes) were unique to the pan-cancer matrix. Calibration analyses were performed for each immune category and pooled curves were plotted (**Supplemental Figure S8**).

#### 3.2.1 Balanced accuracy and precision hold-out test results for Group 1

For immune label categories derived from the pan cancer signature matrix, SVM models predicted overall immune score, CD4 T cells, CD8 T cells, T cells, M0 macrophages, DC and NK cells with balanced accuracies of 0.50, 1.00, 0.50, 1.00, 0.50, 1.00, 1.00, and precision of 0.94, 1.00, 0.88, 1.00, 0.94, 1.00, 1.00, respectively (**Supplemental Table S3a**). Ensemble models predicted these categories with balanced accuracies of 0.47, 1.00, 0.36, 1.00, 0.94, 1.00, 1.00, and precision of 0.94, 1.00, 0.85, 1.00, 1.00, 1.00, 1.00, 1.00, 1.00, respectively. Imbalanced test data labels for T reg cells, TAMs, M1 and M2 macrophages, resulted in metrics with “non evaluable” or “unstable” results (described in detail below).

For immune label categories derived from the glioblastoma signature matrix, SVM models predicted overall immune score, T cells, TAMs, microglia, DC and NK cells with balanced accuracies of 0.50, 0.50, 0.50, 0.50, 0.50, 1.00, and precision of 0.18, 0.35, 0.12, 0.94, 0.65, 1.00, respectively (**Supplemental Table S3a**). Ensemble models predicted the same balanced accuracy and precision.

### 3.2.2 Balanced accuracy and precision hold-out test results for Group 2

For immune label categories derived from the pan cancer signature matrix, SVM models predicted overall immune score, T reg cells, T cells, M1, M2 and M0 macrophages, TAMs, DC and NK cells with balanced accuracies of 0.50, 0.57, 0.63, 1.00, 1.00, 0.50, 0.50, 0.50, 1.00, and precision of 0.53, 0.38, 0.22, 1.00, 1.00, 0.67, 0.44, 0.97, 1.00, respectively (**Supplemental Table S3b**). Ensemble models predicted these categories with balanced accuracies of 0.50, 0.58, 0.34, 1.00, 1.00, 0.50, 0.50, 0.50, 1.0, and precision of 0.53, 0.41, 0.06, 1.00, 1.00, 0.67, 0.44, 0.97, 1.00, respectively. Imbalanced test data labels for CD4 and CD8 T cells resulted in the metrics with "non evaluable" results (described in detail below in 3.2.5).

For immune label categories derived from the glioblastoma signature matrix, SVM models predicted overall immune score, T cells, TAMs, microglia, DC and NK cells with balanced accuracies of 1.00, 0.48, 1.00, 0.50, 0.50, 0.97, and precision of 1.00, 0.61, 1.00, 0.69, 0.47, 1.00, respectively (**Supplemental Table S3b)**. Ensemble models predicted these categories with balanced accuracies of 1.00, 0.45, 1.00, 0.50, 0.52, 1.00, and precision of 1.00, 0.60, 1.00, 0.69, 0.48, 1.00, respectively.

### 3.2.3 Balanced accuracy and precision hold-out test results for Group 3

For immune label categories derived from the pan cancer signature matrix, SVM models predicted CD4 T cells, CD8 T cells, T cells, M1 and M0 macrophages, DC and NK cells with balanced accuracies of 0.50, 0.50, 1.00, 1.00, 0.47, 0.50, 0.50, and precision of 0.79, 0.10, 1.00, 1.00, 0.61, 0.35, 0.59, respectively (**Supplemental Table S3c**). Ensemble models predicted these categories with balanced accuracies of 0.50, 0.50, 1.00, 0.21, 0.58, 0.50, 0.50 and precision of 0.79, 0.10, 1.00, 1.00, 1.00, 0.35, 0.59. Imbalanced test data labels for overall immune scores, T reg cells, TAMs, and M2 macrophages resulted in the metrics with "non evaluable" or "unstable" results (described in detail below in 3.2.5).

For immune label categories derived from the glioblastoma signature matrix, SVM models predicted overall immune score, T cells, microglia, DC and NK cell scores with balanced accuracies of 1.00, 0.48, 0.50, 0.50, 0.50, and precision of 1.00, 0.79, 0.41, 0.86, 0.72, respectively (**Supplemental Table S3c**). Ensemble models predicted overall immune score, T cells, microglia, DC and NK cell scores with balanced accuracies of 1.00, 0.50, 0.50, 0.50, 0.50, and precision of 1.00, 0.79, 0.41, 0.86, 0.72, respectively. Imbalanced test data labels for TAMs scores resulted in the metrics with "non evaluable" results (described in detail below in 3.2.5).

#### 3.2.4 MCC hold-out test results

A moderately positive MCC was obtained for macrophage M0 subtype with ENS_pan model (0.39) in Group 1 held-out results and a weak positive MCC (0.27) for macrophage M0 in Group 3 held-out results (**Table 2**). Similarly, a weak positive signal was obtained for T regulatory cells (SVM_pan model, MCC =0.23) and T cells (SVM_pan model, MCC= 0.18) in group 2 held-out results. MCC results were impacted for some immune categories in each held-out groups with imbalanced test gaussian labels data leading to "non evaluable" or "unstable" results (described in detail below in 3.2.5).

### 3.2.5 Predicting sparse immune labels across held-out datasets

In the evaluation across these three independent held-out datasets (IvyGAP, TCGA-GBM, CPTAC), some of the gaussian-derived immune labels produced only one class (all positive or all negative) and metrics (precision, recall, balanced accuracy and MCC) were mathematically "undefined". Following scikit-learn convention, these metrics collapsed and were interpreted as "non evaluable". In addition, for some of the gaussian-derived immune labels, one class was represented by only a single case. The metrics here were computable but "unstable" as the values depended entirely on the prediction of that one patient score. These limitations were attributed to extreme class imbalance of gaussian-derived immune labels in the held-out test data leading to metric artifacts rather than model failure.

### 3.2.6 Model comparison: bootstrapped resampling

As a separate post hoc experiment to determine if performance accuracy was superior in one model type over another, we applied paired bootstrapped resampling with 10,000 iterations for each immune category, and compared both the precision, balanced accuracy and MCC of the models (SVM_pan versus ENS_pan; SVM_gbm versus ENS_gbm, where pan = pan-cancer signature matrix, gbm = glioblastoma signature matrix, SVM = support vector machine, ENS = ensemble). In the 5 immune categories where there were all 4 models for the same immune category, 6 model pair combinations were determined, and a multiple comparison adjustment was applied. In terms of balanced accuracy, only 4 comparisons showed superiority of one model over another ($p < 0.05$, raw CI) which persisted after multiple comparison adjustment (**Supplemental Table S4**). Here, SVM_pan was shown to be the superior model for T cells, and both glioblastoma models were equally superior for overall immune score. Otherwise, all models were similar in performance although ENS_pan models tended to perform better for macrophage M0 cells scores for precision, balanced accuracy and MCC (all $p < 0.08$).

**3.3 Model selection:**

Overall, ENS_pan models for macrophage M0 scores were the only models demonstrating reasonable stability and robustness after conservative and relevant analyses from three independent experiments using different datasets **(Table 2)**. SVM_pan models showed promise for classification of some scores associated with T cells, but are not stable after rigorous analyses **(Supplemental Table S4)**. SHAP analysis of the macrophage M0 signature radiomic features were performed (**Fig. 2c,d**, **Supplemental Figure S9**).

**4. Discussion:**

**Summary of Findings**

We curated and harmonized the largest open-access IDH-wt glioblastoma dataset containing matched MRI and transcriptomic information. Using pre-operative MR images to identify immune signatures derived from transcriptomic data, we developed a radiogenomic multi-modality machine learning architecture (PRECISE-GBM) (**Fig. 1c**). We performed a robust analytical validation analysis with immune scores classified across three independent experiments, focusing on the meaningful performance metrics of precision and balanced accuracy, and the application of a conservative correlation analysis. Class imbalance almost certainly led to exclusion of multiple models - the consequence of testing diverse hold out datasets. Nonetheless, ensemble models for M0 macrophages gave reasonable stability and robustness and can be considered for future use. SVM models for T cells scores showed some promise, but were not stable after rigorous analyses.

**4.1 Study explanations and relevance**

This study is the first of its kind studying IDH-wt glioblastoma utilizing WHO 2021 classification[10] for immune stratification of patients with development of family of radiogenomic machine learning models[37]. Curation of the largest open access dataset of 176 IDH-wt glioblastoma patients with harmonization across datasets is a major strength.

This study utilizes deep learning based, pretrained, auto-segmentation techniques for glioblastoma tumors on the pre-operative MRI images encompassing all four commonly utilized sequences namely T1, T1 CE, T2 and FLAIR as a part of widely accepted brain tumor protocol for diagnostic imaging. The study also uses the richness of the pixel-based data of radiomics extracted from different regions of the tumor namely enhancing tumor, necrotic core and non-enhancing edema portion giving rise to radiomic signature of each of these areas separately. Utilization of harmonization techniques both for radiomics (neuroCombat) and genomics data (Combat) in the same study are also unique to this field. In terms of immune cell signatures of the TME, we derived these from glioblastoma tumor cells alone as well as from various other cancer cells (pan-cancer). The combination of these two-ground truth immune signature matrices provided us with a broad view of the innate and adaptive immune TME encompassing resident (microglia) and recruited (macrophages, lymphocytes), and immunosuppressive (Tregs, M2) and pro-inflammatory (M1, CD8) profiles.

We have built a pipeline for future radiogenomic studies that is robust in reproducibility, interpretability, and trustworthiness. The proof-of-concept architecture is a combination of PCA as a first pass filter, GMM for immune classifier label distribution, LASSO-CV for feature selection, and SVM along with ensemble models for classification with cross validation and – where there are multiple models for the same immune label – a bootstrapped sampling technique to determine if one model is better than the others.

Our findings from the radiogenomic analyses identify the macrophage M0 signature as the only immune related feature that could be reliably extracted from the MRI images. The pre-operative identification of the M0-related imaging biomarker has potential clinical implications for neo-adjuvant immunotherapy. The current experimental macrophage related targets show a promising approach in targeting this macrophage subtype[61-65]. Individual therapies targeting macrophages have demonstrated that there are plausible rationales to reprogram the TME in pre-clinical and early clinical studies[66-73]. Targeted macrophage therapies (CSF-1/1R inhibitors: pexidartitinib)[74] have been shown to block CSF1R signaling in macrophages tackling its overexpression associated with poor prognosis. Toll-like receptor (TLR) agonists[75], oncolytic viral therapies[76,77], immune checkpoint inhibitors and pre-clinical cytokine therapies (IFN-Y, GM-CSF, IL-12)[78,79] have also been shown to polarize these M0 macrophages to M1 phenotype producing anti-tumor effects. Anti-CD47 therapy (CD47-SIRPα blockade- magrolimab)[80] have been shown to remove the “don’t eat me” signal on tumor cell facilitating phagocytosis of the tumor cells by macrophages and polarize M0 towards M1 subtypes that produce anti-tumor effects.

Box 2: Conceptual radiogenomic enrichment strategies

- In terms of the application of radiogenomic enrichment, biomarker predictions might complement stratification of patients for various immunotherapies in future clinical trial settings (Fig. 4) by providing a non-invasive framework to characterize the immune cell state of the glioblastoma TME prior to confirmation by biopsy.

- Integration of predicted immune scores could enable classification of tumors into stratification categories of broadly inflamed ("immune-hot"), immunosuppressed ("immune-cold") or mixed phenotypes based on the relative abundance of cytotoxic lymphocytes (e.g. CD8 T cells, NK cells) and immunosuppressive populations such as TAMs, M2 macrophage, microglia, T regulatory cells[16-20,24]. Such immune phenotypes have been shown to influence responsiveness to immunotherapy and other immune-modulating strategies in preclinical and clinical studies of glioblastoma[8,16-25].

- Enriching patient cohorts could be considered for emerging immunotherapy strategies in clinical trial settings prior to biopsy confirmation and final trial stratification. For example, this paradigm could be employed in tumors with relatively inflamed immune profiles which may be more amenable to immune checkpoint blockade strategies. Similarly, tumors with dominant macrophage or microglial signature may require combination approaches aimed at reprogramming the immunosuppressive TME.

### 4.5 Limitations of the study:

Several areas were identified as likely to impact on the results of our study.

Firstly, the absence of a curated radiogenomic dataset was a limitation for our project. This gap necessitated the development of a curated radiogenomic dataset utilized in this study. To address this limitation and facilitate future research, we provide open-access to this curated radiogenomic dataset for IDH-wt glioblastoma. This initiative will support the broader scientific community in advancing radiogenomic research and developing more accurate predictive models. Nonetheless, the availability of matched imaging and genomic datasets was limited, resulting in a relatively small sample size for our study. This constraint has implications for the robustness of our findings and the generalizability of our models. Areas where models struggled to produce accurate predictions are thus a combination of the constraint on the data availability and distribution as well as biological variability of these cells being represented in the TME reflecting intra- and inter-tumoral heterogeneity. Larger datasets would be beneficial in overcoming bias and variance limitations, allowing for more comprehensive training and improved validation accuracy. Future research should aim to incorporate more extensive harmonized datasets to ultimately enhance the reliability and applicability of the models developed.

Secondly, our study is a retrospective multicenter investigation utilizing datasets from multiple sites, each employing different MRI machines and protocols. While we employed a common registration space and the neuroCombat tool to mitigate these discrepancies, inherent variations remain. These variations can introduce biases and affect the consistency of our results. As an alternative to building larger datasets to overcoming bias and variance limitations, prospective studies with standardized imaging protocols across sites could help to minimize these differences and improve the uniformity of the data.

Similarly, differences in genomic data arose from the use of various RNA sequencing and microarray techniques, as well as differing reference genomes between datasets. We obtained gene-level fpkm values and harmonized the RNA sequencing data and microarray data using the Combat tool for all datasets to reduce variability. Addressing these genomic differences remains a challenge and standardized genomic data processing protocols and spatial transcriptomics or single cell sequencing data would benefit validation in future studies.

In summary, it was clear from the outset that variability in the distribution of data across different datasets posed a challenge. Batch effects, stemming from inherent differences in sequencing for genomic data and hardware differences for imaging, posed a significant challenge potentially impacting model performance. We mitigated this by implementing steps for stratification (during the data split for training and testing), normalization, harmonization and cross validation. We utilized batch correction tools embedded in CIBERSORTx[45], neuroCombat, and Combat. Ensemble modeling was a further mitigation during cross validation. Furthermore, GMM provided additional robustness in identifying the distribution of the label signature. However, residual variability impacted the results, and class imbalance almost certainly led to exclusion of multiple models - the consequence of testing diverse hold out datasets. Ongoing refinement of batch correction techniques might improve model accuracy and reliability. Future work on feature engineering and dimensionality reduction should continue too – in particular, multi-task learning is likely to help predict labels jointly leveraging interdependence.

### 4.2 Current evidence in the field

There is sparse literature on development of radiogenomic models and various methods have been employed for training of the data and validation of the results[37]. SVM and ensemble models have been shown to be powerful in performing binary classification and have been utilized previously to perform classification tasks based on genomic and radiomic data[53-57]. Ensemble models utilize the strength of multiple algorithms, capture non-linear patterns, are robust to noise in the data, and are known to outperform single classifier models by reducing risk of under- or over-fitting patterns in high-dimensional data such as in radiogenomics[53-57]. Bioinformatics-based digital cytometry has been employed to derive immune-related signatures in various cancers including glioblastoma and has been shown to be a robust and reproducible methodology[32,45,46]. Regarding classifier data distribution, gaussian mixture models have been shown to independently predict the gaussian distribution of the data in imbalanced datasets and thus aid in radiogenomic model development[47,48]. Furthermore, LASSO-CV methodology is a commonly utilized technique for feature selection to obtain robust features that can be then utilized downstream in the radiogenomic model training[81-83]. Likewise, PCA[84,85] can identify outliers following linear dimensionality reduction and acts as a first-pass filter to reduce noise in linearly separable variance patterns of the radiomics data. Cross-cohort held out strategies have been shown to be useful in high-dimensional biomedical data and data originating from multiple cohorts providing a realistic estimate of the performance of the model on unseen real-world datasets; making it ideal for multi-source radiogenomics data as compared to other cross validation techniques [52,86-88].

Apart from SVM, various other machine learning models such as gradient boost, logistic regression, KNN, random forest, symbolic regression have been utilized in literature to develop radiogenomic models[32]. Bootstrapped resampling techniques have been widely utilized to deal with model selection when multiple model comparisons are required which we encountered in this study[59,60,89].

In terms of the very few radiogenomic models of the immune system in glioblastoma, previous studies have employed ADC values, nCBV values, or radiomic features (VASARI, texture, shape, histogram, and wavelet)[37]. However, only two studies have used internal hold-out datasets for analytical biomarker validation, and none have used external hold-out datasets to validate the trained biomarker - inferences regarding generalizability are therefore severely limited[37]. In contrast we performed three independent experiments with three external hold-out datasets.

**4.6 Future implications**

The development and analytical validation of these radiogenomic models serves both as a proof-of-concept and a plausible pathway for macrophage M0 immune signatures to be used as biomarkers. PRECISE-GBM is intended for baseline, pre-operative immune characterization in newly diagnosed IDH-wt glioblastoma. Future work could explore recurrent disease which would likely require a change in training data followed by further external validation. To enable translation, two key steps are needed.

First, the PRECISE-GBM pipeline requires a prerequisite step to ensure that inputs represent patients with IDH-wt glioblastoma only. Accurate identification and exclusion of glioblastoma mimics such as grade 3 and 4 astrocytoma, grade 3 oligodendroglioma, and metastatic tumors are essential for clinical deployment. Several published models appear to be able to perform this prerequisite step[90-92].

Second, the results of this study require confirmation in a prospective clinical validation study to strengthen the findings. We recommend that for initial clinical validation, the macrophage subtype

M0 signature is used to prospectively predict response in neo-adjuvant immunotherapy clinical trials in shadow mode, with the prediction unblinded after the trial is complete.

In terms of model development to increase predictive power, future research could incorporate other novel immune markers such as tumor mutation burden, neoantigen markers, and T cell receptor markers. It is also conceivable that future developments might also incorporate liquid biopsy specimens in multi-modal models.

Ultimately, should a range of radiogenomic models be clinically validated in the future, a panel of pre-trained models could be chosen by an agentic algorithm to produce individualized immune profiles for patients with IDH-wt glioblastoma. These profiles could serve as personalized biomarkers for personalized neo-adjuvant immunotherapy trials.

**Supplemental data:** Supplementary File (Figures and tables)

**Ethical statement:** All the anonymized retrospective data were obtained from open access portals with appropriate permission. Data curation and analysis was performed in accordance with the Declaration of Helsinki. No explicit ethics approval was necessary for this study.

**Funding**

TCB supported by the Wellcome Trust [WT 203148/Z/16/Z]. PG is supported by King's College London postgraduate research (PGR) international studentship. MM is supported by Chronic Disease Research Foundation (CDRF). Open access article processing charges (APC) are funded by King's College London.

**Disclosure**

A part of the data has been presented as oral presentation at British Neuro-oncology Society (BNOS) conference 2024, European association of Neuro-oncology (EANO) congress 2024, Society of British Neurological Surgeons (SBNS) spring meeting 2025 and MICCAI 2025.

**Acknowledgements**

We would like to thank ARTORG Center for Biomedical Engineering Research, University of Bern for facilitating MRI auto-segmentation pipeline; Leeds GBMdeconvoluteR Team for providing glioblastoma signature matrix; NEWMAN lab at Stanford University for support on bioinformatics deconvolution pipeline; Beijing Tiantan Hospital and CGGA Team for facilitating analysis of CGGA data and BMEIS Team for facilitating hardware for this project. We would like to thank Alysha Chelliah, Benjamin Jackson, Golestan Karami, Siddharth Agarwal, Liane Dos Santos Canas in BMEIS for helping during the project. Illustrative figures were created in Biorender.com.

**CRediT authorship contribution statement**

PG: Writing – review & editing, Writing – original draft, Visualization, Resources, Methodology, Investigation, Formal analysis, Data curation, Validation; PG, MM, TCB: Conceptualization; JL, LY: Dataset collection; PG, MM, TCB: Supervision, Writing – review & editing; all authors have read and approved the final manuscript.

**Consent for publication**

The data utilized in this study was exclusively open-access and anonymized. No explicit consent was required for analysis of the open-access data.

**Code and data availability Statement**

Code for pre-processing the radiogenomic data and model architecture are available in https://github.com/prazg/PRECISE_GBM. Imaging dataset links are available in Supplementary Table S3.

**Declaration of competing interest**

Authors declare no known competing interests.

**References:**

1. Sharma P, Aaroe A, Liang J, Puduvalli VK. Tumor microenvironment in glioblastoma: Current and emerging concepts. Neurooncol Adv. 2023;5(1):vdad009. Published 2023 Feb 23. doi:10.1093/noajnl/vdad009
2. Habashy KJ, Mansour R, Moussalem C, Sawaya R, Massaad MJ. Challenges in glioblastoma immunotherapy: mechanisms of resistance and therapeutic approaches to overcome them. Br J Cancer. 2022;127(6):976-987. doi:10.1038/s41416-022-01864-w
3. Sampson JH, Gunn MD, Fecci PE, Ashley DM. Brain immunology and immunotherapy in brain tumours. Nat Rev Cancer. 2020;20(1):12-25. doi:10.1038/s41568-019-0224-7
4. Wang M, Zhou Z, Wang X, Zhang C, Jiang X. Natural killer cell awakening: unleash cancer-immunity cycle against glioblastoma. Cell Death Dis. 2022;13(7):588. Published 2022 Jul 8. doi:10.1038/s41419-022-05041-y
5. Yang M, Oh IY, Mahanty A, Jin WL, Yoo JS. Immunotherapy for Glioblastoma: Current State, Challenges, and Future Perspectives. Cancers (Basel). 2020;12(9):2334. Published 2020 Aug 19. doi:10.3390/cancers12092334

6. Rocha Pinheiro SL, Lemos FFB, Marques HS, Silva Luz M, de Oliveira Silva LG, Faria Souza Mendes Dos Santos C, da Costa Evangelista K, Calmon MS, Sande Loureiro M, Freire de Melo F. Immunotherapy in glioblastoma treatment: Current state and future prospects. World J Clin Oncol. 2023 Apr 24;14(4):138-159. doi: 10.5306/wjco.v14.i4.138. PMID: 37124134; PMCID: PMC10134201.
7. Liu Y, Zhou F, Ali H et al. Immunotherapy for glioblastoma: current state, challenges, and future perspectives. Cell Mol Immunol 2024;**21**:1354–1375. doi:10.1038/s41423-024-01226-x
8. Long GV, Shklovskaya E, Satgunaseelan L, et al. Neoadjuvant triplet immune checkpoint blockade in newly diagnosed glioblastoma. Nat Med. Published online February 27, 2025. doi:10.1038/s41591-025-03512-1
9. Ellenbogen Y, Zadeh G. A new paradigm for immunotherapy in glioblastoma. Nat Med. Published online March 31, 2025. doi:10.1038/s41591-025-03607-9
10. Louis DN, Perry A, Wesseling P, et al. The 2021 WHO Classification of Tumors of the Central Nervous System: a summary. Neuro Oncol. 2021;23(8):1231-1251. doi:10.1093/neuonc/noab106
11. Bouffet E, Larouche V, Campbell BB, et al. Immune Checkpoint Inhibition for Hypermutant Glioblastoma Multiforme Resulting From Germline Biallelic Mismatch Repair Deficiency. J Clin Oncol. 2016;34(19):2206-2211. doi:10.1200/JCO.2016.66.6552
12. Johanns TM, Miller CA, Dorward IG, et al. Immunogenomics of Hypermutated Glioblastoma: A Patient with Germline POLE Deficiency Treated with Checkpoint Blockade Immunotherapy. Cancer Discov. 2016;6(11):1230-1236. doi:10.1158/2159-8290.CD-16-0575

13. Nabian N, Ghalehtaki R, Zeinalizadeh M, Balaña C, Jablonska PA. State of the neoadjuvant therapy for glioblastoma multiforme-Where do we stand? Neurooncol Adv. 2024 Mar 5;6(1):vdae028. doi: 10.1093/noajnl/vdae028. PMID: 38560349; PMCID: PMC10981465.
14. Reardon DA, Brandes AA, Omuro A, et al. Effect of Nivolumab vs Bevacizumab in Patients With Recurrent Glioblastoma: The CheckMate 143 Phase 3 Randomized Clinical Trial. JAMA Oncol. 2020;6(7):1003-1010. doi:10.1001/jamaoncol.2020.1024
15. Basak U, Sarkar T, Mukherjee S, Chakraborty S, Dutta A, Dutta S, Nayak D, Kaushik S, Das T, Sa G. Tumor-associated macrophages: an effective player of the tumor microenvironment. Front Immunol. 2023 Nov 16;14:1295257. doi: 10.3389/fimmu.2023.1295257. PMID: 38035101; PMCID: PMC10687432.
16. Reardon DA, Gokhale PC, Klein SR, et al. Glioblastoma Eradication Following Immune Checkpoint Blockade in an Orthotopic, Immunocompetent Model. Cancer Immunol Res. 2016;4(2):124-135. doi:10.1158/2326-6066.CIR-15-0151
17. Bloch O, Crane CA, Kaur R, Safaee M, Rutkowski MJ, Parsa AT. Gliomas promote immunosuppression through induction of B7-H1 expression in tumor-associated macrophages. Clin Cancer Res. 2013;19(12):3165-3175. doi:10.1158/1078-0432.CCR-12-3314
18. Wainwright DA, Chang AL, Dey M, et al. Durable therapeutic efficacy utilizing combinatorial blockade against IDO, CTLA-4, and PD-L1 in mice with brain tumors. Clin Cancer Res. 2014;20(20):5290-5301. doi:10.1158/1078-0432.CCR-14-0514
19. Zeng J, See AP, Phallen J, et al. Anti-PD-1 blockade and stereotactic radiation produce long-term survival in mice with intracranial gliomas. Int J Radiat Oncol Biol Phys. 2013;86(2):343-349. doi:10.1016/j.ijrobp.2012.12.025

20. De Martino M, Daviaud C, Lira MC, Hernandez-Zirofsky K, Vanpouille-Box C. Dual blockade of PD-1 and CTLA-4 generates long-lasting immunity against irradiated glioblastoma. Cancer Lett. 2025;628:217856. doi:10.1016/j.canlet.2025.217856

21. Boutilier AJ, Elsawa SF. Macrophage Polarization States in the Tumor Microenvironment. Int J Mol Sci. 2021 Jun 29;22(13):6995. doi: 10.3390/ijms22136995. PMID: 34209703; PMCID: PMC8268869.

22. Christofides A, Strauss L, Yeo A, Cao C, Charest A, Boussiotis VA. The complex role of tumor-infiltrating macrophages. *Nat Immunol*. 2022;23(8):1148-1156. doi:10.1038/s41590-022-01267-2

23. ClinicalTrials.gov. https://clinicaltrials.gov/study/NCT06816927

24. Cloughesy TF, Mochizuki AY, Orpilla JR, et al. Neoadjuvant anti-PD-1 immunotherapy promotes a survival benefit with intratumoral and systemic immune responses in recurrent glioblastoma. Nat Med. 2019;25(3):477-486. doi:10.1038/s41591-018-0337-7

25. McFaline-Figueroa JR, Sun L, Youssef GC, et al. Neoadjuvant anti-PD1 immunotherapy for surgically accessible recurrent glioblastoma: clinical and molecular outcomes of a stage 2 single-arm expansion cohort. Nat Commun. 2024;15(1):10757. Published 2024 Dec 30. doi:10.1038/s41467-024-54326-7

26. Gabrusiewicz K, Rodriguez B, Wei J, et al. Glioblastoma-infiltrated innate immune cells resemble M0 macrophage phenotype. JCI Insight. 2016;1(2):e85841. doi:10.1172/jci.insight.85841

27. Huang L, Wang Z, Chang Y, Wang K, Kang X, Huang R, Zhang Y, Chen J, Zeng F, Wu F, Zhao Z, Li G, Huang H, Jiang T, Hu H. EFEMP2 indicates assembly of M0 macrophage and more

malignant phenotypes of glioma. Aging (Albany NY). 2020 May 12;12(9):8397-8412. doi: 10.18632/aging.103147. Epub 2020 May 12. PMID: 32396873; PMCID: PMC7244085.

28. Moreno-Sanchez PM, Rezaeipour M, Inderberg EM, Platten M, Golebiewska A. Immunosuppressive mechanisms and therapeutic interventions shaping glioblastoma immunity. Nat Cancer. 2026;7(1):29-42. doi:10.1038/s43018-025-01097-9
29. Singh G, Singh A, Bae J, et al. -New frontiers in domain-inspired radiomics and radiogenomics: increasing role of molecular diagnostics in CNS tumor classification and grading following WHO CNS-5 updates. Cancer Imaging. 2024;24(1):133. Published 2024 Oct 7. doi:10.1186/s40644-024-00769-6
30. Chen D, Zhang R, Huang X, et al. MRI-derived radiomics assessing tumor-infiltrating macrophages enable prediction of immune-phenotype, immunotherapy response and survival in glioma. Biomark Res. 2024;12(1):14. Published 2024 Jan 31. doi:10.1186/s40364-024-00560-6
31. Ellingson BM. Radiogenomics and imaging phenotypes in glioblastoma: novel observations and correlation with molecular characteristics. Curr Neurol Neurosci Rep. 2015;15(1):506. doi:10.1007/s11910-014-0506-0
32. Liu D, Chen J, Ge H, et al. Radiogenomics to characterize the immune-related prognostic signature associated with biological functions in glioblastoma. Eur Radiol. 2023;33(1):209-220. doi:10.1007/s00330-022-09012-x
33. Khalili N, Kazerooni AF, Familiar A, et al. Radiomics for characterization of the glioma immune microenvironment. NPJ Precis Oncol. 2023;7(1):59. Published 2023 Jun 19. doi:10.1038/s41698-023-00413-9

34. Fathi Kazerooni A, Bagley SJ, Akbari H, Saxena S, Bagheri S, Guo J, Chawla S, Nabavizadeh A, Mohan S, Bakas S, Davatzikos C, Nasrallah MP. Applications of Radiomics and Radiogenomics in High-Grade Gliomas in the Era of Precision Medicine. Cancers (Basel). 2021 Nov 25;13(23):5921. doi: 10.3390/cancers13235921. PMID: 34885031; PMCID: PMC8656630.
35. Chaddad A, Kucharczyk MJ, Daniel P, Sabri S, Jean-Claude BJ, Niazi T, Abdulkarim B. Radiomics in Glioblastoma: Current Status and Challenges Facing Clinical Implementation. Front Oncol. 2019 May 21;9:374. doi: 10.3389/fonc.2019.00374. PMID: 31165039; PMCID: PMC6536622.
36. Aftab K, Aamir FB, Mallick S, et al. Radiomics for precision medicine in glioblastoma. J Neurooncol. 2022;156(2):217-231. doi:10.1007/s11060-021-03933-1
37. Ghimire P, Kinnersley B, Golestan K, Arumugam P, Houlston R, Ashkan K, Modat M, Booth CT. Radiogenomic biomarkers for immunotherapy in glioblastoma: a systematic review of magnetic resonance imaging studies, Neuro-Oncology Advances, 2024;, vdae055, https://doi.org/10.1093/noajnl/vdae055
38. Cagney DN, Sul J, Huang RY, Ligon KL, Wen PY, Alexander BM. The FDA NIH Biomarkers, EndpointS, and other Tools (BEST) resource in neuro-oncology. *Neuro Oncol*. 2018;20(9):1162-1172. doi:10.1093/neuonc/nox242
39. Scarpace L, Mikkelsen, T, Cha, S, et al. The Cancer Genome Atlas Glioblastoma Multiforme Collection (TCGA-GBM) (Version 5) [Data set]. The Cancer Imaging Archive.2016
40. National Cancer Institute Clinical Proteomic Tumor Analysis Consortium (CPTAC). The Clinical Proteomic Tumor Analysis Consortium Glioblastoma Multiforme Collection (CPTAC-GBM) (Version 16) [dataset]. The Cancer Imaging Archive. 2018 https://doi.org/10.7937/K9/TCIA.2018.3RJE41Q1

41. Shah N, Feng X, Lankerovich M, Puchalski RB, Keogh B. Data from Ivy Glioblastoma Atlas Project (IvyGAP) [Data set]. The Cancer Imaging Archive. 2016. https://doi.org/10.7937/K9/TCIA.2016.XLwaN6nL
42. Scarpace L, Flanders AE, Jain R, Mikkelsen T, Andrews, DW. Data From REMBRANDT [Data set]. The Cancer Imaging Archive. 2019. https://doi.org/10.7937/K9/TCIA.2015.588OZUZB
43. Zhao Z, Zhang KN, Wang Q, et al. Chinese Glioma Genome Atlas (CGGA): A Comprehensive Resource with Functional Genomic Data from Chinese Glioma Patients. Genomics, Proteomics & Bioinformatics. 2021 Feb;19(1):1-12
44. Conte GM, Weston AD, Vogelsang DC, et al. Generative Adversarial Networks to Synthesize Missing T1 and FLAIR MRI Sequences for Use in a Multisequence Brain Tumor Segmentation Model. Radiology.2021;299(2):313-323. doi:10.1148/radiol.2021203786
45. Newman AM, Steen CB, Liu CL, et al. Determining cell type abundance and expression from bulk tissues with digital cytometry. Nat Biotechnol. 2019;37(7):773-782. doi:10.1038/s41587-019-0114-2
46. Ajaib S, Lodha D, Pollock S, et al. GBMdeconvoluteR accurately infers proportions of neoplastic and immune cell populations from bulk glioblastoma transcriptomics data. Neuro Oncol. 2023;25(7):1236-1248. doi:10.1093/neuonc/noad021
47. Goodfellow, I., Bengio, Y., & Courville, A. (2016). Deep learning. MIT Press.

48. Naglik, I., Lango, M. GMMSampling: a new model-based, data difficulty-driven resampling method for multi-class imbalanced data. Mach Learn 2023;113:5183-5202. doi:10.1007/s10994-023-06416-8

49. NITRC: DeepBraTumIA: Tool/Resource Info. Retrieved April 4, 2024, from https://www.nitrc.org/projects/deepbratumia/

50. Suter Y, Knecht U, Valenzuela W, Notter M, Hewer E, Schucht P, Wiest R, Reyes M. The LUMIERE dataset: Longitudinal Glioblastoma MRI with expert RANO evaluation. Sci Data. 2022 Dec 15;9(1):768. doi: 10.1038/s41597-022-01881-7. PMID: 36522344; PMCID: PMC9755255.

51. van Griethuysen JJM, Fedorov A, Parmar C, et al. Computational Radiomics System to Decode the Radiographic Phenotype. Cancer Res. 2017;77(21):e104-e107. doi:10.1158/0008-5472.CAN-17-0339

52. Fortin JP, Cullen N, Sheline YI, et al. Harmonization of cortical thickness measurements across scanners and sites. Neuroimage. 2018;167:104-120. doi:10.1016/j.neuroimage.2017.11.024

53. Cervantes J, García-Lamont F, Rodríguez-Mazahua L, López A. A comprehensive survey on support vector machine classification: Applications, challenges and trends. Neurocomputing. 2020;408:189-215. doi:10.1016/j.neucom.2019.10.118

54. Xi YB, Guo F, Xu ZL, et al. Radiomics signature: A potential biomarker for the prediction of MGMT promoter methylation in glioblastoma. J Magn Reson Imaging. 2018;47(5):1380-1387. doi:10.1002/jmri.25860

55. Zhang X, Yan LF, Hu YC, et al. Optimizing a machine learning based glioma grading system using multi-parametric MRI histogram and texture features. Oncotarget. 2017;8(29):47816-47830. doi:10.18632/oncotarget.18001

56. Tian Q, Yan LF, Zhang X, et al. Radiomics strategy for glioma grading using texture features from multiparametric MRI. J Magn Reson Imaging. 2018;48(6):1518-1528. doi:10.1002/jmri.26010

57. Artzi M, Bressler I, Ben Bashat D. Differentiation between glioblastoma, brain metastasis and subtypes using radiomics analysis. J Magn Reson Imaging. 2019;50(2):519-528. doi:10.1002/jmri.26643

58. Chicco D, Jurman G. The advantages of the Matthews correlation coefficient (MCC) over F1 score and accuracy in binary classification evaluation. BMC Genomics. 2020 Jan 2;21(1):6. doi: 10.1186/s12864-019-6413-7.

59. Tsamardinos I, Greasidou E, Borboudakis G. Bootstrapping the out-of-sample predictions for efficient and accurate cross-validation. Mach Learn. 2018;107(12):1895-1922. doi: 10.1007/s10994-018-5714-4. Epub 2018 May 9.

60. Lubke GH, Campbell I, McArtor D, Miller P, Luningham J, van den Berg SM. Assessing Model Selection Uncertainty Using a Bootstrap Approach: An update. Struct Equ Modeling. 2017;24(2):230-245. doi: 10.1080/10705511.2016.1252265. Epub 2016 Dec 5. PMID: 28652682; PMCID: PMC5482523.

61. Khan M, Huang X, Ye X, et al. Necroptosis-based glioblastoma prognostic subtypes: implications for TME remodeling and therapy response. Ann Med. 2024;56(1):2405079. doi:10.1080/07853890.2024.2405079

62. Li H, Tang Y, Hua L, et al. A Systematic Pan-Cancer Analysis of MEIS1 in Human Tumors as Prognostic Biomarker and Immunotherapy Target. J Clin Med. 2023;12(4):1646. Published 2023 Feb 18. doi:10.3390/jcm12041646

63. Yuan F, Cong Z, Cai X, et al. BACH1 as a potential target for immunotherapy in glioblastomas. Int Immunopharmacol. 2022;103:108451. doi:10.1016/j.intimp.2021.108451

64. Chang Y, Li G, Zhai Y, et al. Redox Regulator GLRX Is Associated With Tumor Immunity in Glioma. Front Immunol. 2020;11:580934. Published 2020 Nov 30. doi:10.3389/fimmu.2020.580934

65. Ye J, Yang Y, Dong W, et al. Drug-free mannosylated liposomes inhibit tumor growth by promoting the polarization of tumor-associated macrophages. Int J Nanomedicine. 2019;14:3203-3220. Published 2019 May 2. doi:10.2147/IJN.S207589

66. Rao R, Han R, Ogurek S, et al. Glioblastoma genetic drivers dictate the function of tumor-associated macrophages/microglia and responses to CSF1R inhibition. Neuro Oncol. 2022;24(4):584-597. doi:10.1093/neuonc/noab228

67. Quail DF, Bowman RL, Akkari L, et al. The tumor microenvironment underlies acquired resistance to CSF-1R inhibition in gliomas. Science. 2016;352(6288):aad3018. doi:10.1126/science.aad3018

68. Azambuja JH, Schuh RS, Michels LR, et al. Blockade of CD73 delays glioblastoma growth by modulating the immune environment. Cancer Immunol Immunother. 2020;69(9):1801-1812. doi:10.1007/s00262-020-02569-w

69. Felsenstein M, Blank A, Bungert AD, et al. CCR2 of Tumor Microenvironmental Cells Is a Relevant Modulator of Glioma Biology. Cancers (Basel). 2020;12(7):1882. Published 2020 Jul 13. doi:10.3390/cancers12071882

70. Zheng Z, Zhang J, Jiang J, et al. Remodeling tumor immune microenvironment (TIME) for glioma therapy using multi-targeting liposomal codelivery. J Immunother Cancer. 2020;8(2):e000207. doi:10.1136/jitc-2019-000207

71. Yang F, He Z, Duan H, et al. Synergistic immunotherapy of glioblastoma by dual targeting of IL-6 and CD40. Nat Commun. 2021;12(1):3424. Published 2021 Jun 8. doi:10.1038/s41467-021-23832-3

72. Zhang X, Chen L, Dang WQ, et al. CCL8 secreted by tumor-associated macrophages promotes invasion and stemness of glioblastoma cells via ERK1/2 signaling. Lab Invest. 2020;100(4):619-629. doi:10.1038/s41374-019-0345-3

73. Tang F, Wang Y, Zeng Y, Xiao A, Tong A, Xu J. Tumor-associated macrophage-related strategies for glioma immunotherapy. *NPJ Precis Oncol*. 2023;7(1):78. Published 2023 Aug 19. doi:10.1038/s41698-023-00431-7

74. Mendez JS, Cohen AL, Eckenstein M, et al. Phase 1b/2 study of orally administered pexidartinib in combination with radiation therapy and temozolomide in patients with newly diagnosed glioblastoma. *Neurooncol Adv*. 2024;6(1):vdae202. Published 2024 Nov 22. doi:10.1093/noajnl/vdae202

75. Tiwari RK, Singh S, Gupta CL, et al. Repolarization of glioblastoma macrophage cells using non-agonistic Dectin-1 ligand encapsulating TLR-9 agonist: plausible role in regenerative medicine against brain tumor. *Int J Neurosci*. 2021;131(6):591-598. doi:10.1080/00207454.2020.1750393

76. Nassiri F, Patil V, Yefet LS, et al. Oncolytic DNX-2401 virotherapy plus pembrolizumab in recurrent glioblastoma: a phase 1/2 trial. *Nat Med*. 2023;29(6):1370-1378. doi:10.1038/s41591-023-02347-y

77. Yang Y, Brown MC, Zhang G, Stevenson K, Mohme M, Kornahrens R, Bigner DD, Ashley DM, López GY, Gromeier M. Polio virotherapy targets the malignant glioma myeloid infiltrate with diffuse microglia activation engulfing the CNS. Neuro Oncol. 2023 Sep 5;25(9):1631-1643. doi: 10.1093/neuonc/noad052. PMID: 36864784; PMCID: PMC10479910.

78. Chiocca EA, Yu JS, Lukas RV, et al. Regulatable interleukin-12 gene therapy in patients with recurrent high-grade glioma: Results of a phase 1 trial. *Sci Transl Med*. 2019;11(505):eaaw5680. doi:10.1126/scitranslmed.aaw5680
79. Banerjee K, Núñez FJ, Haase S, McClellan BL, Faisal SM, Carney SV, Yu J, Alghamri MS, Asad AS, Candia AJN, Varela ML, Candolfi M, Lowenstein PR, Castro MG. Current Approaches for Glioma Gene Therapy and Virotherapy. Front Mol Neurosci. 2021 Mar 11;14:621831. doi: 10.3389/fnmol.2021.621831. PMID: 33790740; PMCID: PMC8006286.
80. *ClinicalTrials.gov*. Retrieved August 28, 2025, from https://clinicaltrials.gov/study/NCT05169944
81. Kawahara D, Tang X, Lee CK, Nagata Y, Watanabe Y. Predicting the Local Response of Metastatic Brain Tumor to Gamma Knife Radiosurgery by Radiomics With a Machine Learning Method. Front Oncol. 2021;10:569461. Published 2021 Jan 11. doi:10.3389/fonc.2020.569461
82. Park JE, Kim HS, Jo Y, et al. Radiomics prognostication model in glioblastoma using diffusion- and perfusion-weighted MRI. Sci Rep. 2020;10(1):4250. Published 2020 Mar 6. doi:10.1038/s41598-020-61178-w
83. Chang E, Joel MZ, Chang HY, et al. Comparison of radiomic feature aggregation methods for patients with multiple tumors. Sci Rep. 2021;11(1):9758. Published 2021 May 7. doi:10.1038/s41598-021-89114-6
84. Jolliffe, I.T., Cadima, J. Principal component analysis: A review and recent developments. Philos. Trans. R. Soc. A Math. Eng. Sci. 2016;374: 20150202.

85. Lambin P, Leijenaar RTH, Deist TM, et al. Radiomics: the bridge between medical imaging and personalized medicine. Nat Rev Clin Oncol. 2017;14(12):749-762. doi:10.1038/nrclinonc.2017.141

86. Geroldinger A, Lusa L, Nold M, Heinze G. Leave-one-out cross-validation, penalization, and differential bias of some prediction model performance measures-a simulation study. Diagn Progn Res. 2023 May 2;7(1):9. doi: 10.1186/s41512-023-00146-0. PMID: 37127679; PMCID: PMC10152625

87. Yamashita A, Yahata N, Itahashi T, et al. Harmonization of resting-state functional MRI data across multiple imaging sites via the separation of site differences into sampling bias and measurement bias. PLoS Biol. 2019;17(4):e3000042. Published 2019 Apr 18. doi:10.1371/journal.pbio.3000042

88. Orlhac F, Boughdad S, Philippe C, et al. A Postreconstruction Harmonization Method for Multicenter Radiomic Studies in PET. J Nucl Med. 2018;59(8):1321-1328. doi:10.2967/jnumed.117.199935

89. Philipp Koehn. Statistical Significance Tests for Machine Translation Evaluation. In Proceedings of the 2004 Conference on Empirical Methods in Natural Language Processing, 2004;388–395.

90. Calabrese E, Villanueva-Meyer JE, Cha S. A fully automated artificial intelligence method for non-invasive, imaging-based identification of genetic alterations in glioblastomas. Sci Rep. 2020 Jul 16;10(1):11852. doi: 10.1038/s41598-020-68857-8. PMID: 32678261; PMCID: PMC7366666.

91. Bathla G, Dhruba DD, Soni N, et al. AI-based classification of three common malignant tumors in neuro-oncology: A multi-institutional comparison of machine learning and deep learning methods. J Neuroradiol. 2024;51(3):258-264. doi:10.1016/j.neurad.2023.08.007

92. Shin I, Kim H, Ahn SS, Sohn B, Bae S, Park JE, Kim HS, Lee SK. Development and Validation of a Deep Learning-Based Model to Distinguish Glioblastoma from Solitary Brain Metastasis Using Conventional MR Images. AJNR Am J Neuroradiol. 2021 May;42(5):838-844. doi: 10.3174/ajnr.A7003. Epub 2021 Mar 18. PMID: 33737268; PMCID: PMC8115383.

## Figures

**Figure 1.** PRECISE-GBM workflow

**Figure 2:** Non-invasive radiomic signature of macrophage subtype M0 immune phenotype.

**Figure 3**: Forest plot comparing models across immune categories and signature matrices using cross-cohort held-out strategy.

**Figure 4:** Pipeline for integrating PRECISE-GBM models into clinical practice

## Tables

**Table 1:** Baseline characteristics of IDH-wt glioblastoma patients within the matched transcriptome and imaging datasets that were combined into Groups and used for training

**Table 2**: Preferred model for immune category.

**Figure 1.** PRECISE-GBM workflow

a. Schematic showing how the matched imaging and transcriptome datasets were used to derive immune-related score predictions. Five datasets contained matched imaging and transcriptomic data. Two signature matrices were utilized during deconvolution to estimate immune related score classifiers. The datasets were grouped into three groups with three holdout datasets. Three trained group models were obtained with SVM alone, as well as an ensemble model incorporating SVM, RF and GB models, to predict different immune binary labels.

b. Synthetic image workflow showing pix2pix inspired[44] GAN model training for generating missing T2 FLAIR (FLAIR) images from T2-weighted (T2) images using UCSF PDGM dataset (490 patients); generation of gFLAIR sequences using trained generator; and segmentation and extraction of radiomic features for training of PRECISE-GBM models using gFLAIR images.

c. PRECISE-GBM architecture. MRI images were segmented and radiomics extracted, which then underwent LASSO feature selection. The transcriptomic data underwent deconvolution to estimate immune TME and generate immune-related scores. These scores were binarized using GMM and the binary labels used to inform radiomic feature selection (dotted arrow), and ultimately for training of classifier models. The trained model was used for prediction of these scores in held-out test datasets; there was never any data leakage.

mRNA: messenger ribonucleic acid; PCA: principal component analysis; LASSO-CV: Least absolute shrinkage and selection operator cross validation; GMM: gaussian mixture model; SVM: support vector machine; MRI: magnetic resonance imaging; TME: tumor microenvironment; T1: T1-weighted MRI sequence, T2: T2-weighted MRI sequence; FLAIR: fluid attenuated inversion recovery T2 MRI sequence; T1 CE: T1 MRI sequence with contrast enhancement; CV: cross validation; SVM: support vector machine; RF: random forest; GB: histogram-based gradient boost; PRECISE-GAN: Production of REconstructed Complete Imaging using Synthetic Engineering for Glioma through Generative Adversarial Network; gFLAIR: generated FLAIR; CGGA: Chinese Glioma Genome Atlas; UCSF PDGM: University of California San Francisco Preoperative Diffuse Glioma MRI; GAN: Generative Adversarial network; PRECISE-GBM: Predictive Radiomics for Evaluation of Cancer Immune SignaturE in GlioBlastoMa (Created in Biorender.com)

**Figure 2**. Non-invasive radiomic signature of macrophage subtype M0 immune phenotype.

a. Explainability SHAP analysis for the PRECISE-GBM radiomic signature for macrophage subtype M0.

The analysis demonstrated that predictions were primarily driven by higher-order texture features. Across scenarios, the most influential features belonged to the first order, GLSZM, GLCM and GLDM families. These features exerted moderate and directionally consistent contributions to model output with higher effect size in Model 1 scenario but markedly reduced in Model 2 scenario suggesting preserved biological signal with lower confidence in Model 2 due to cohort-specific variability. In Model 3 scenario, the same feature classes reappear but their effects are bidirectional and less monotonic.

Across all three scenarios, the model consistently relies on spatial heterogeneity metrics (GLCM cluster prominence, GLSZM zone variance), large scale structural organization (large area emphasis, large dependence emphasis) and global intensity distribution (energy, total energy) reflecting spatially extensive, heterogenous, tumor regions, possibly consistent with an immune-naïve or transitional myeloid environment. No single radiomic feature dominated predictions across all patients, indicating stable and distributed decision-making with evidence of robust generalization. The SHAP values are close to zero demonstrating the model is stable and not overly sensitive to that feature for most patients.

Each row is a radiomic feature, ranked from top to bottom by mean absolute SHAP value (global importance). Each dot is one patient in the held-out test set; x-axis (SHAP value) represents contribution of that feature to the model output – positive value pushes the prediction toward

"immune-high/ class 1"; negative value pushes towards "immune-low/class 0"; red dots – high feature value; blue dots-low feature value.

b. Kaplan-Meier overall survival (OS) curve stratified by predicted macrophage M0 immune classes. OS data were plotted for the predicted high (n = 54) and low (n = 26) macrophage subtype M0 signatures based on the PRECISE-GBM classifier. Whilst our aim was to build diagnostic biomarkers for patient stratification for immunotherapy treatment we noted that patients predicted as M0-high demonstrated longer OS compared to those predicted as M0-low (log-rank test p = 0.04), demonstrating that the M0 signature is also a prognostic biomarker.

c. Overlap of radiomic features across MRI sequences for the macrophage M0 immune signature.

Venn diagram showing the distribution of radiomic features contributing to the macrophage M0 immune classifier across FLAIR, T1, T1 CE and T2 sequences. The large intersection (feature count = 767) reflects features shared across multiple sequences, while sequence-specific regions highlight unique radiomic features derived from individual MRI sequences, demonstrating complementary imaging information captured by different sequences.

d. Overlap of radiomic features across tumor regions for the macrophage M0 immune signature.

Venn diagram illustrating the distribution of radiomics features associated with the macrophage M0 immune classifier across three tumor regions: 1- necrotic core, 2- enhancing tumor and 3- edema region. The diagram demonstrates both shared and region-specific radiomics features, indicating that while large number of features (feature count = 1134) are common across multiple tumor regions, additional features uniquely arise from specific spatial tumor compartments.

SHAP: Shapley Additive exPlanations; PRECISE-GBM: Predictive Radiomics for Evaluation of Cancer Immune SignaturE in GlioBlastoMa ; GLSZM: gray level size zone matrix ; GLCM: Gray

Level Co-occurrence Matrix ; GLDM: Gray Level Dependence Matrix; ENS: ensemble; T1: T1-weighted MRI sequence, T2: T2-weighted MRI sequence; FLAIR: fluid attenuated inversion recovery T2 MRI sequence; T1 CE: T1 MRI sequence with contrast enhancement; IDH: isocitrate dehydrogenase."

**Figure 3**: Forest plot comparing models across immune categories and signature matrices using cross-cohort held-out strategy. The plot displays mean performance metrics (precision, balanced accuracy and MCC) for SVM and ensemble models, with bars representing the range of results across three independent held-out test sets (model 1, 2 and 3). Immune categories are shown in the y-axis and metric values on the corresponding x -axis. Two panels depict the results for glioblastoma signature and pan-cancer signature matrices. Points are color-coded by classifier type (SVM-blue, ensemble-red). Full results table for each group is available in Supplemental Table S3.

TAMs: tumor associated macrophages; DC: dendritic cells; NK cells: natural killer cells; SVM: support vector machine ; MCC: Matthews correlation coefficient

**Figure 4**: Conceptual pipeline for integrating PRECISE-GBM models into future clinical trials. The strategy for the binary predictions is based on biologically-informed immune cell functions in the TME, grouping the binary classification results into TME categories. Trialists can use this information to enrich groups of patients who could be recruited into trials for reprogramming of the TME or combination therapies after category confirmation and stratification by biopsy. (PRECISE-GBM model: Predictive Radiomics for Evaluation of Cancer Immune SignaturE in Glioblastoma model; TME: tumor microenvironment) (currently restricted to tumor associated macrophage subtype M0) (Created in Biorender.com).

**Table 1:** Baseline characteristics of IDH-wt glioblastoma patients within the matched transcriptome and imaging datasets that were combined into Groups and used for training. (Group 1: CGGA, TCGA-GBM, CPTAC, REMBRANDT; Group 2: CGGA, CPTAC, IvyGAP, REMBRANDT; Group 3: CGGA, TCGA-GBM, IvyGAP, REMBRANDT). IvyGAP was held out for Group 1; TCGA-GBM was held out for Group 2 and CPTAC was held out for Group 3. There was no significant difference between the available baseline characteristics of the three Groups used for training (all ANOVA except Chi squared test for M:F ratio; $p$-value < 0.05 considered statistically significant).

OS: overall survival; SD: standard deviation.

**Table 2**: Preferred model for immune category. Binary gaussian class of immune categories namely macrophage M0 and T cells were predicted from the radiomic models (ensemble for macrophage M0 and SVM for T cells) accurately in multiple held-out datasets consistently supported by the metrics results. Macrophage M0 category was supported by precision, BA, moderately positive MCC results – and the particular model favored after bootstrapping in terms of BA and MCC. T cells category was supported by precision, BA, weakly positive MCC results – and the particular model selected by bootstrapping in terms of BA. Full bootstrapping results are available in **Supplemental Table S4**.

SVM: support vector machine; ENS: ensemble model; MCC: Matthews correlation coefficient; BA: Balanced accuracy

**Table 1:** Baseline characteristics of IDH-wt glioblastoma patients within the matched transcriptome and imaging datasets that were combined into Groups and used for training. (Group 1: CGGA, TCGA-GBM, CPTAC, REMBRANDT; Group 2: CGGA, CPTAC, IvyGAP, REMBRANDT; Group 3: CGGA, TCGA-GBM, IvyGAP, REMBRANDT). IvyGAP was held out for Group 1; TCGA-GBM was held out for Group 2 and CPTAC was held out for Group 3. There was no significant difference between the available baseline characteristics of the three Groups used for training (all ANOVA except Chi squared test for M:F ratio; p-value < 0.05 considered statistically significant).

| Variable | PRECISE-GBM curated radiogenomic matched dataset groups | | | p value | Test statistic |
|---|---|---|---|---|---|
| | Group 1 | Group 2 | Group 3 | | |
| Sample size | 155 | 141 | 143 | | |
| Mean age ± SD (years) | 56.58 ± 13.08 | 56.60 ± 13.12 | 55.44 ± 12.40 | 0.109 | 2.59 |
| M:F ratio | 1.33:1 | 1.13.1 | 1.21:1 | 0.444 | 0.59 |
| Mean OS +/- SD (years) | 1.56 ± 1.75 | 1.55 ± 1.73 | 1.69 ± 1.78 | 0.983 | 0.00047 |
| Tumor volume ($cm^3$) | 115.07 ± 57.06 | 116.03 ± 55.89 | 114.36 ± 55.60 | 0.438 | 0.60 |

OS: overall survival; SD: standard deviation.

**Table 2**: Preferred model for immune category. Binary gaussian class of immune categories namely macrophage M0 and T cells were predicted from the radiomic models (ensemble for macrophage M0 and SVM for T cells) accurately in multiple held-out datasets consistently supported by the metrics results. Macrophage M0 category was supported by precision, BA, moderately positive MCC results – and the particular model favored after bootstrapping in terms of BA and MCC. T cells category was supported by precision, BA, weakly positive MCC results – and the particular model selected by bootstrapping in terms of BA. Full bootstrapping results are available in **Supplemental Table S4**.

SVM: support vector machine; ENS: ensemble model; MCC: Matthews correlation coefficient; BA: Balanced accuracy

| Immune Category | Preferred Signature Matrix | Preferred Model | Metric results (Group 1, Group 2, Group 3 mean results; supplemented by bootstrapping) |
|---|---|---|---|
| Macrophages M0 | Pan-cancer | ENS | Mean BA 0.67 (0.50-0.94);<br>Mean Precision 0.89 (0.67- 1.00);<br>Moderately positive MCC (mean MCC 0.22);<br>Favored on BA, precision and MCC bootstrapping |
| T cells | Pan-cancer | SVM | Mean BA 0.88 (0.63-1.00);<br>Mean Precision 0.74 (0.22-1.00);<br>Weakly positive MCC (mean MCC 0.06),<br>Statistical significance on BA bootstrapping;<br>Favored on precision and MCC bootstrapping |

a

PRECISE-GBM dataset (n=176)

CV

Prediction of binary labels

Cross-cohort held-out strategy

Stratified 5-fold internal hyperparameter tuning

Immune score

Overall T cells Score

CD4 T cells Score

CD8 T cells Score

T Regulatory cells Score

Overall TAMs Score

Macrophage M0 Score

Macrophage M1 Score

Macrophage M2 Score

Microglia Score

Dendritic cells (DC) Score

Natural Killer cells (NK) Score

Pan cancer signature matrix

Immune Matrices

Glioblastoma signature matrix

Classifiers

PRECISE-GBM
SVM and Ensemble models

b

PRECISE-GAN model

Input

Target

T2

FLAIR

Generator

Output

gFLAIR

FLAIR

Discriminator

PRECISE-GAN model training

T2 volume

Trained Generator

gFLAIR volume

Generation of missing FLAIR sequences

DeepBratumIA model

CGGA MRI sequences

Auto-segmentation of tumour regions

Sequence-specific Radiomic Features

Generation of Radiomic Features

c

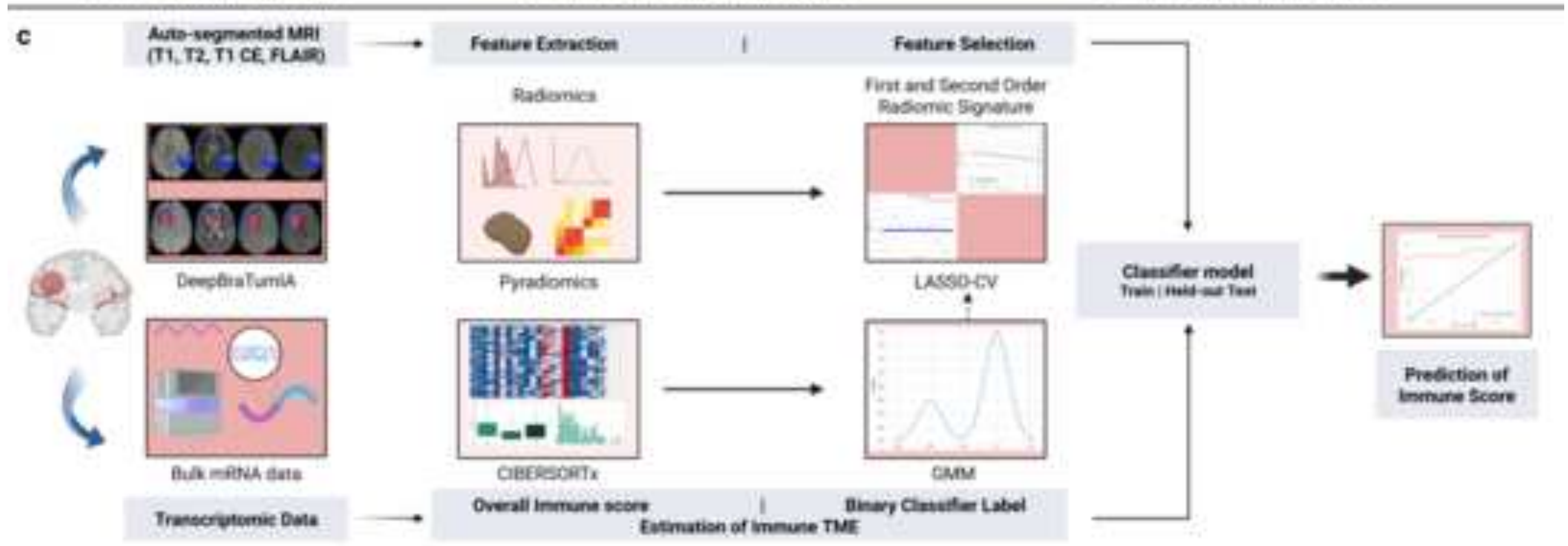

Figure 2

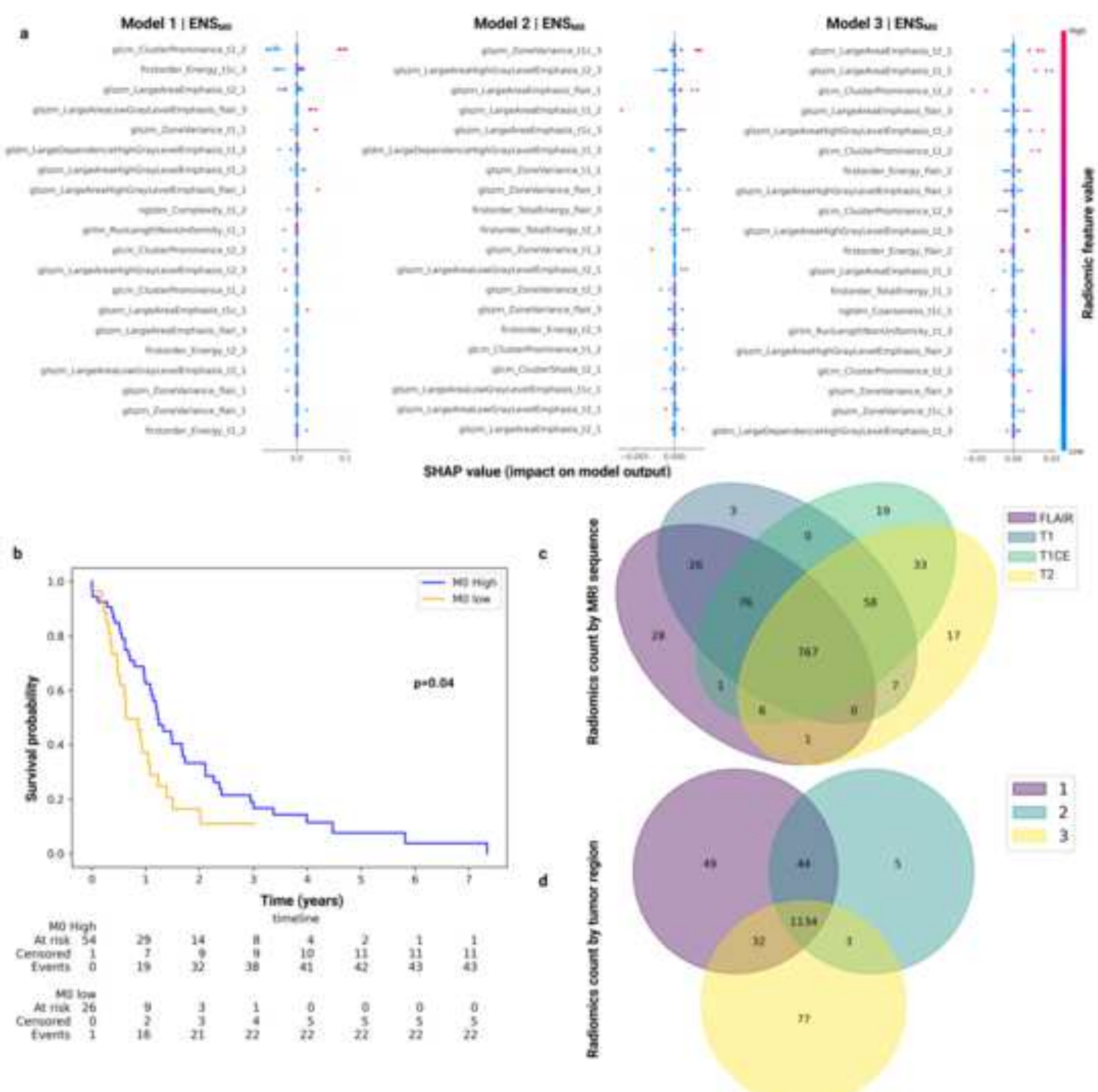
a
Model 1 | ENS
Model 2 | ENS
Model 3 | ENS
SHAP value (impact on model output)
Radiomic feature value
High
Low
b
MO High
MO low
p=0.04
Survival probability
Time (years)
timeline
MO High
At risk 54 29 14 8 4 2 1 1
Censored 1 7 9 9 10 11 11 11
Events 0 19 32 38 41 42 43 43
MO low
At risk 26 9 3 1 0 0 0 0
Censored 0 2 3 4 5 5 5 5
Events 1 16 21 22 22 22 22 22
c
Radiomics count by MRI sequence
FLAIR
T1
T1CE
T2
767
d
Radiomics count by tumor region
1
2
3
49
44
5
1134
32
1
77

Figure 3 Click here to access/download;Figure;Figure3_PRECISE.tiff

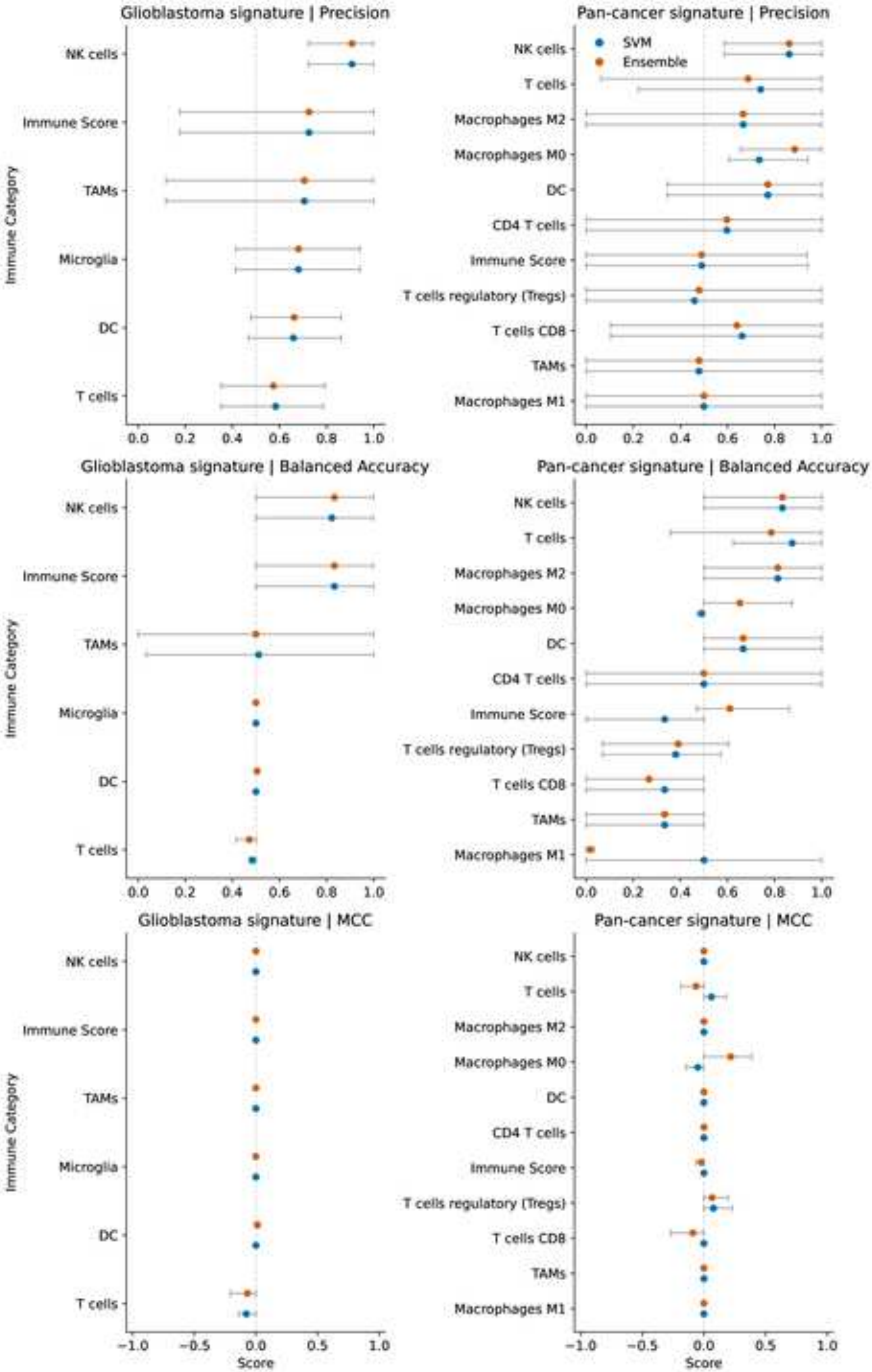

Figure 4 Click here to access/download;Figure;Figure_4_PRECISE.tiff

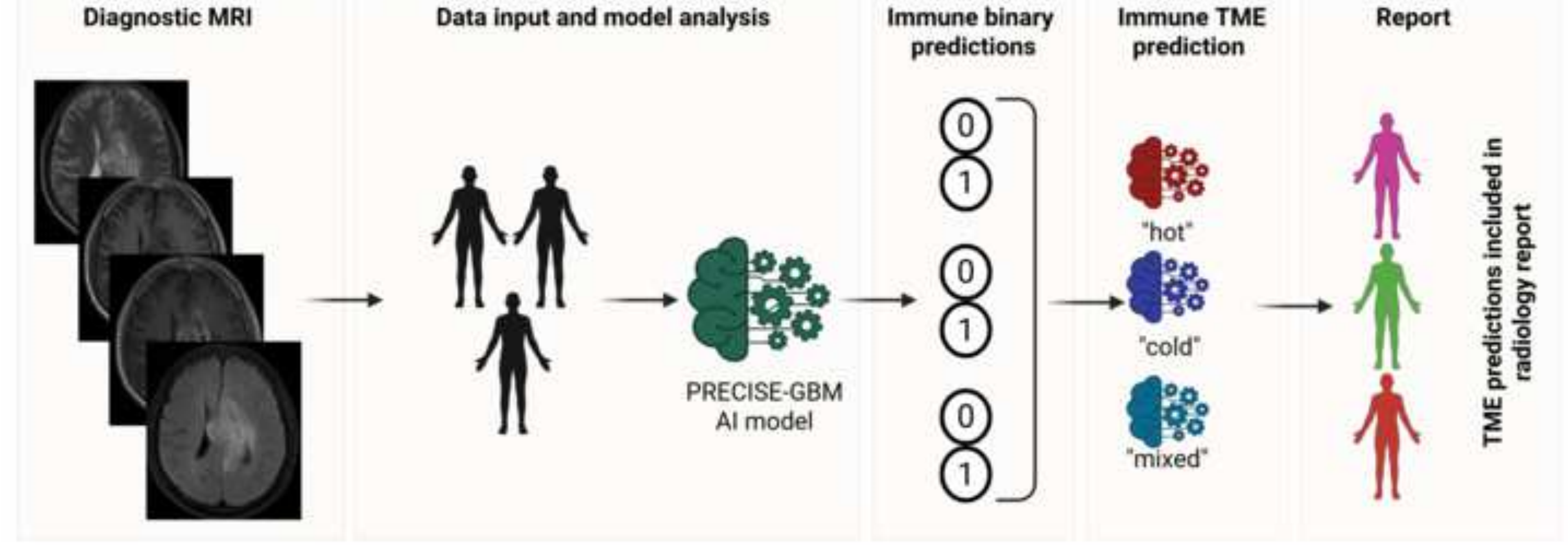